\documentclass{article}
\usepackage{graphicx} % Required for inserting images
\usepackage{amsmath}
\usepackage{booktabs}
\usepackage{bm}
\usepackage{hyperref}
\usepackage{tabularx}
\usepackage{xcolor}
\usepackage{ulem}
\usepackage{enumitem}
\usepackage{cite}
\usepackage{url}
\title{\textbf{Piccolo2: General Text Embedding with Multi-task Hybrid Loss Training }}
\date{}
\author{
\\
\parbox{\linewidth}{
\text{Junqin Huang, Zhongjie Hu, Zihao Jing, Mengya Gao, Yichao Wu} 
}\\   
\\
\textbf{SenseTime Research} \\
}

\begin{document}

\maketitle
\begin{abstract}
In this report, we introduce Piccolo2, an embedding model that surpasses other models in the comprehensive evaluation over 6 tasks on CMTEB \cite{xiao2023c} benchmark, setting a new state-of-the-art. Piccolo2 primarily leverages an efficient multi-task hybrid loss training approach, effectively harnessing textual data and labels from diverse downstream tasks. In addition, Piccolo2 scales up the embedding dimension and uses MRL\cite{kusupati2022matryoshka} training to support more flexible vector dimensions. The latest information of piccolo models can be accessed via: \href{https://huggingface.co/sensenova/}{piccolo}

\end{abstract}

\section{Introduction}\label{sec:intro}
 Text embedding models play a pivotal role in natural language processing and machine learning. By encoding texts into structured numerical representations, known as text embeddings, these models encapsulate semantic and contextual information of words, phrases, or entire documents within a dense, low-dimensional vector space\cite{reimers2019sentence}. Such embeddings are indispensable for various downstream NLP tasks, including classification, clustering, retrieval, and sentence similarity. \\
 
 Contrastive learning stands out as the most effective technique for training text embedding models \cite{gao2021simcse}. It presents text semantic representations by minimizing the distance between positive pairs and maximizing the distance between negative pairs. Beyond its application in natural language processing (NLP), contrastive learning also proves pivotal in visual\cite{He_2020_CVPR}\cite{chen2020simple} and multi-modal\cite{radford2021learning} representation learning. Recent advanced text embedding works \cite{xiao2023c}\cite{wang2022text}\cite{li2023towards} primarily rely on a two-stage pretrain-finetune pipeline to acquire general text embedding models. Pre-training utilizes weakly supervised data sourced from large-scale crawling efforts, while fine-tuning involves refining the model with high-quality text pairs obtained through data mining or manual annotation. \\

In this paper, we introduce Piccolo2 and propose a multi-task hybrid training method to better utilize textual data and labels from different granularities (e.g. usually the labels of STS\cite{nlizh} tasks are more fine-grained than retrieval\cite{xie2023t2ranking} tasks). In addition to training methods, high-quality data has long been recognized as an important component in embedding training\cite{wang2022text}\cite{lee2024gecko}. Piccolo2 also devise a data synthetic framework and a hard negative mining approach to continually augment high-quality datasets. We demonstrate our training approach in Section \ref{sec:train}, the composition and source of the data in Section \ref{sec:data}, and our experimental results in Section \ref{sec:exp}. 
\\

\section{Training Details}\label{sec:train}
\subsection{Multi-Task Hybrid Loss Training}\label{sec:multilossft}
Previous training processes for embedding models are mostly relied on the standard InfoNCE loss \cite{gutmann2010noise} with in-batch negative samples, which typically achieves robust representations through the utilization of large batch size and a great number of negative samples. However, standard InfoNCE cannot resolve all situations in the present landscape of downstream tasks for embedding models. For example, STS tasks usually achieve inferior results by training with InfoNCE loss. Furthermore, classification tasks and clustering tasks have also not been utilized in the training of general embedding models. Hence, we employ a multi-task hybrid loss training method to Piccolo2, which utilize various training loss for different downstream task and prove its superior performance. In the following parts, we introduce the loss functions of our hybrid training methods in detail. \\

\subsubsection{Retrieval and Reranking Loss}
For retrieval and reranking tasks\cite{xie2023t2ranking}, we follow previous works \cite{wang2022text}\cite{li2023towards} and use the standard InfoNCE loss with in-batch negative.

\begin{equation} \label{infonce}
    L_{re} = - \frac{1}{n} \sum_i \log \frac{\text{e}^{s_(q,d^+)/\tau}}{\text{e}^{s_(q,d^+)/\tau}+\sum_j \text{e}^{s_(q,d^-)/\tau} }
\end{equation}

where $s(q,d)$ is a scoring function between query $q$ and passage $p$, often defined as the cosine similarity and $\tau$ is the scale temperature.

\subsubsection{STS and PairClassification Loss}
Most previous studies\cite{wang2022text}\cite{xiao2023c} also use InfoNCE loss to optimize tasks on natural language inference (NLI) datasets, where contradictory sentences are usually treated as hard negatives, while entailment or neutral sentences are treated as positives. However, since the original STS and pair-classification tasks usually have more fine-grained labels (such as score values), converting them into triplets inevitably leads to information loss, consequently yielding inferior results. Therefore, we directly employ the cosent loss function \cite{cosent}, a ranking loss function specifically designed for the text pairs with fine-grained labels.

\begin{equation} \label{cosent}
    \mathcal{L}_{sts} = \mathrm{log} \left ( 1 + \sum _{s(\mathbf{x}_i, \mathbf{x}_j) > s(\mathbf{x}_m, \mathbf{x}_n)} e^{\frac{\mathrm{cos}(\mathbf{x}_m, \mathbf{x}_n) - \mathrm{cos}(\mathbf{x}_i, \mathbf{x}_j)}{\tau}} \right )
\end{equation}
where $\tau$ is the scale temperature, cos(·) is the cosine similarity function, and s(u, v) is the
similarity between u and v.

% For most STS datasets in NLI tasks, the input composition is usually {'text', 'text pair', 'score'}. The score is the similarity of the pair, usually 0-5. This type of data usually cannot be used directly through infoNCE loss for finetune. In the initial experiment, we used a simple way to unify them, that is, identifying samples with score larger than threshold as positive samples, constructing data {text, text pos}, and text neg is empty. Samples with score smaller than threshold are considered negative samples and are constructed into data {text, text pos, text neg}, where text and text pos are the same text, usually the threshold is set to 2.5.

\subsubsection{Classification and Clustering Loss} \label{losscls}
For classification and clustering tasks, we leverage the SFR embedding method \cite{SFRAIResearch2024} to seamlessly integrate data into training process. To be more specific, we reformat classification and clustering datasets into contrastive triplets. For example, in a classification task with 10 categories, each input text $x$ is treated with its target label $y^+$ as the positive pair, while the remaining 9 labels $y^-$ serve as negative pairs. This approach is similarly applied to clustering tasks. This differs from the approach mentioned in \cite{lee2024gecko}, where documents sharing the same label are considered as positive pairs, and those with different labels are considered as negative pairs. In contrast, pairing documents with their corresponding labels is more intuitive.

\begin{equation} \label{infonce_cls}
    L_{cls} = - \frac{1}{n} \sum_i \log \frac{\text{e}^{s_(x,y^+)/\tau}}{\text{e}^{s_(x,y^+)/\tau}+\sum_j \text{e}^{s_(x,y^-)/\tau} }
\end{equation}
Formula \ref{infonce} and Formula \ref{infonce_cls} share the same structure. However, it's important to stress that, for $L_{cls}$, in-batch negatives are no longer used due to the fact that it can easily lead to conflict training targets. As a result, only label negatives are used in $L_{cls}$. \\

\subsubsection{Multi-Task Hybrid Loss} \label{hybrid loss}

During training, we blend the aforementioned three loss functions and final loss function is formulated as follows:
\begin{equation} \label{mthl}
L = \left\{
\begin{array}{ll}
    L_{cls} & \text{if task is classification or clustering} , \\
    L_{sts} & \text{if task is sts or pair-classification}, \\
    L_{re} & \text{if task is retrieval or reranking},
\end{array}
\right.
\end{equation}

\subsection{Dimension Scaling up and MRL Training}
\subsubsection{Dimension Scaling up}
Inspired by OpenAI's text-embedding-v3\cite{text-embedding-v3} we scale up the embedding vector dimension from 768 to 1792 to increase the model capacity, compared with Piccolo1\cite{piccolo}. During the fine-tuning stage, We directly add a learnable linear layer to the last layer of BERT, with a size of $(origin\_dim, scale\_dim)$. The operation can be formulated as:
\begin{equation} \label{scale_dim}
    emb = Linear(pool(BERT(text)))
\end{equation}

\subsubsection{MRL Training}
Matryoshka Representation Learning \cite{kusupati2022matryoshka} introduces a novel approach for training embedding models with flexible dimension lengths, ensuring the utility of embeddings even after dimensionality reduction. This technique not only enhances the speed of processing but also significantly reduces storage requirements with slight performance drop. This has proven to be very effective in text-embedding v3 released by OpenAI\cite{text-embedding-v3} and has been adopted by many advanced embeddings works such as \cite{emb2024mxbai}\cite{lee2024gecko} \cite{stella}. In the training of Piccolo2, we also employ the MRL approach and validate its effectiveness.

\section{Datasets}\label{sec:data}
Piccolo2 is trained on a diverse range of tasks. We collect open-source data and employ a helpful data synthetic pipeline to generate more training samples. Hard negative mining is also applied on dataset for retrieval tasks.

\subsection{Datasets Synthetic Pipeline} \label{data_syn_pp}
Recently, a lot of work has focused on generating large-scale sample pairs through GPT4 to eliminate the need for complicated manual annotation \cite{wang2023improving}\cite{lee2024gecko}. This is a very useful method, especially in some special scenarios where data is scarce. We follow previous work \cite{wang2023improving} by creating approximately 200k retrieval datasets and incorporating them into our training datasets. For further details, please refer to the Appendix.
\\

% In order to be able to generate more diverse data, we adopted a two-stage approach to data generation. In the first phase, we let GPT4 randomly generate a large number of topic phrases in various domains to ensure the diversity of the data. To get high quality topic phrases, we prepare a list containing several retrieval topic phrase examples, and each time we generate a new topic phrase, we will randomly select two examples from the list as reference examples and add them to the prompt. In the second phase, we generate text pairs containing query, positive text and negative text based on the large number of topic phrases obtained from the previous generation through a specially designed prompt. In order to enrich the dataset, we generate text pairs 10 times for each topic phrase, and for each text pair, we set the corresponding random parameter in the prompt to make the text pairs obtained each time completely different. For example, for query generation, we randomly set its word count, clarity, etc., and randomly set the length and comprehensibility of positive text and negative text. It should be mentioned that for the definition of negative text, we require that the result generated by GPT4 should contain some useful information in the query, but its usefulness and comprehensiveness should not exceed that of the positive text. In the appendix, we give the prompts we used.
\subsection{Dataset Details}
For retrieval tasks, we collect data from MMarcoRetrieval\cite{bonifacio2021mmarco}, T2Rerieval\cite{xie2023t2ranking}, DuRetrieval\cite{qiu2022dureader_retrieval}, CmedqaRetrieval\cite{8548603},CovidRetrieval\cite{wang2020cord}, and Multi-CPR\cite{long2022multi}. For MMarcoRetrieval, 400k subsets are sampled. Clustering tasks consist of data from Thunews\cite{li2006comparison} and CSL\cite{li2022csl}, with filters applied to exclude development and test sets in the CMTEB clustering framework. In classification, our training dataset utilize datasets from Tnews, Iflytek \cite{xu2020clue}, Multilingual-sentiments\cite{mcauley2013hidden}, JDReview\cite{jdreview}, OnlineShopping, and Waimai\cite{onlineshopping}. For Semantic Textual Similarity (STS) and Pair Classification, training data includes NLI-zh\cite{nlizh}, Afqmc, Qbqtc, Cmnli\cite{xu2020clue} and Ocnli\cite{hu2020ocnli}. Additionally, about 200k short to long retrieval task data are generated through the data synthesis pipeline mentioned in Section \ref{data_syn_pp}. For most of the data sets mentioned here, we use its train split to prevent overlap with the CMTEB test set. We list the composition of the dataset in Table \ref{tab:sup_finetune_data}.

\begin{table}[ht]
\centering
\scalebox{0.95}{\begin{tabular}{lcc}
\hline Task & training data meta & \# Sampled  \\ \hline
STS & {text, text pair, score} &  730k       \\
Pair Classification & {text, text pair, score} & 440k \\ 
Retrieval & {text, pos text, neg text} & 1.1M    \\
Retrieval generated & {text, pos text, neg text} & 200k \\
Clustering & {text, pos label, neg label} & 1M   \\
Classifcation & {text, pos label, neg label} &  220k \\
Total        &  &  $\sim$3.7M       \\ \hline
\end{tabular}}
\caption{Data mixture for supervised fine-tuning.}
\label{tab:sup_finetune_data}
\end{table}

\subsection{Hard Negative Mining}
For each retrieval task, we use piccolo-base-zh\cite{piccolo} to conduct negative sample mining. We randomly select 15 samples from the mining negatives of rank 50 - 100 as the final hard negative samples. We avoid using higher-rank negative samples as their inclusion typically leads to a decline in performance. This is caused by a variety of reasons, such as inaccurate dataset annotation.

\section{Experiments}\label{sec:exp}
\begin{table}[t]
    \small
    \centering
    \resizebox{1.0\linewidth}{!}{%
	\begin{tabular}{lcc|cccccc|c}
		\toprule
		& Dim. & \# Params. & Class. & Cluster. & Pair. & Rerank. & Retr. & STS & Avg. \\
		\midrule
        gte-Qwen1.5-7B-instruct\cite{gte-qwen} & 4,096 & 7B & 73.35 & 67.08 & 88.52 & 66.38 & 70.62 & 62.32 &  69.56 \\
		acge-text-embedding\cite{acge} & 1,792 & 300M & 72.75 & 58.7 & 87.84 & 67.98 & 72.93 & 62.09 &  69.07 \\
		aliyun-text-embedding\cite{open-search} & 1,792 & n/a & 71.74 & 53.75 & 88.1 & 68.27 & 74.41 & 62.46 & 68.71 \\
		stella-mrl-large\cite{stella} & 1,792 & 300M & 71.56 & 54.32 & 88.08 & 68.45 & 73.52 & 62.48  & 68.55 \\
        baichuan-text-embedding\cite{baichuan-embedding} & 1,024 & n/a & 72.84 & 56.88 & 82.32 & 69.67 & 73.12 & 60.07 & 68.34 \\
        \midrule
        \textbf{piccolo2} & 1,792 & 300M & 74.59 & 62.17 & 90.24 & 70 & 74.36 & 63.5 & 70.95 \\
     \bottomrule
	\end{tabular}
}
\caption{Results on CMTEB. We report the average performance on six different tasks: Classification (Class.), Clustering (Cluter.), Pair Classification (Pair.), Reranking (Rerank.), Retrieval (Retr.), and STS. The last column shows the average performance across all datasets from the six tasks.}\label{tab:mteb}
\end{table}
\subsection{Implementation details}
During the training phase, we employ the AdamW\cite{loshchilov2017decoupled} optimizer with an initial learning rate of 1e-5, incorporating cosine decay. The batch size per GPU is set to 256 with only 1 negative sample for each retrieval data point.  The model undergoes training for 2500 steps with a maximum input length limited to 512, and for simplicity we do not include any instructions. For MRL training, we configure the matryoshka representation dimension as 256, 384, 768, 1024, 1536 and 1972. For efficiency, we use mixed precision training, gradient checkpointing and deepspeed ZERO-stage1 \cite{rajbhandari2020zero}. It finally takes 32 A100 GPU and 6 hours for the whole fine-tuning process. It is worth noting that since different tasks (e.g. clustering, retrieval) usually have different number of negatives, we do not use cross device negative because this would be very troublesome to implement in the context of multi loss fine-tuning we mentioned in section \ref{sec:multilossft}.

\subsection{CMTEB evaluation}
MTEB (Massive Text Embedding Benchmark) \cite{muennighoff2022mteb} is a widely used benchmark for evaluating large-scale text embedding tasks. Recently, Xiao et al. \cite{xiao2023c} has introduced CMTEB, which builds upon MTEB by incorporating numerous Chinese datasets, thereby greatly enhancing the evaluation capabilities for Chinese language models. In this report, we conduct experiments on CMTEB, which has 31 datasets spanning across 6 categories: Classification, Clustering, Pair Classification, Rerank, Retrieval and STS. Table \ref{tab:mteb} presents a comparison of our models with others on the CMTEB benchmark. Notably, our best-performing Piccolo2 model surpasses the previous state-of-the-art BERT-based model, acge-text-embeddings, by 1.9 points.

\begin{table}[t]
    \small
    \centering
    \resizebox{1.0\linewidth}{!}{%
	\begin{tabular}{l|cccccc|c}
		\toprule
		& Class. & Cluster. & Pair. & Rerank. & Retrieval & STS & Avg. \\
		\midrule
		  $La : L_{re}$ & 72.85 & 54.72 & 86.14 & 69.71 & 74.05 & 61.88 & 68.75  \\
            $Lb : L_{re}$ + $L_{sts}$ & 73.10 & 53.39 & 90.02 & 69.08 & 74.25 & 63.35 & 69.87  \\
            $Lc : L_{re}$ + $L_{sts}$ + $L_{cls}$ & 74.59 & 62.17 & 90.24 & 70 & 74.36 & 63.5 & 70.95  \\
     \bottomrule
	\end{tabular}
}
\caption{We contrast the effects of employing the hybrid loss against utilizing the InfoNCE loss. The adoption of the hybrid loss has yielded substantial enhancements across clustering, classification, semantic text similarity (STS), and pair classification tasks.}\label{tab:ablation}
\end{table}

\begin{table}[t]
    \small
    \centering
    \resizebox{1.0\linewidth}{!}{%
	\begin{tabular}{l|cccccc|c}
		\toprule
		Eval Dim. & Class. & Cluster. & Pair. & Rerank. & Retrieval & STS & Avg. \\
		\midrule
		1,792 & 74.59 & 62.17 & 90.24 & 70 & 74.36 & 63.5 & 70.95  \\
            1,536 & 74.46 & 62.67 & 90.28 & 69.96 & 74.35 & 63.5 & 70.97  \\
            1,280 & 74.29 & 62.39 & 90.27 & 69.8 & 74.29 & 63.51 & 70.87  \\
            1,024 & 74.05 & 62.27 & 90.3 & 69.93 & 74.27 & 63.52 & 70.8  \\
            768 & 73.79 & 62.29 & 90.3 & 69.62 & 74.21 & 63.52 & 70.69  \\
            512 & 73.36 & 61.85 & 90.17 & 69.37 & 73.87 & 63.47 & 70.41  \\
            256 & 72.74 & 62.24 & 90.1 & 68.89 & 72.88 & 63.4 & 69.99  \\
     \bottomrule
	\end{tabular}
}
\caption{Results on MRL evaluation. We train the model using a 1792-dimensional embedding space and evaluat its performance across various dimensions. Notably, even when the embedding dimension is significantly reduced by roughly eightfold (from 1792 to 256), the model's performance degradation is minimal, at only around 1 point.}\label{tab:mrl}
\end{table}

\subsection{Ablation Study}
\subsubsection{Multi-Task Hybrid Loss tuning}
To demonstrate the effectiveness of hybrid loss tuning compared to directly using InfoNCE for fine-tuning, we conduct a simple ablation experiment. In this experiment, we design three different loss function variants and use the same dataset for training. Our three loss function variants are:

\begin{itemize}[label=\textbullet]
    \item Loss Variant A ($L_{a}$): Only InfoNCE is used for training on the six tasks.
    \item Loss Variant B ($L_{b}$): Cosent loss is used specifically for the STS and Pair Classification tasks, while InfoNCE is used for the rest Retrieval, Classification and Clustering tasks.
    \item Loss Variant C ($L_{c}$): Based on $L_{b}$, it specially utilizes loss function \ref{infonce_cls} for both Clustering and Classification tasks. $L_{c}$ is also our final hybrid loss function \ref{mthl}.
\end{itemize}

% Loss Variant A ($L_{a}$): Only InfoNCE is used for fine-tuning.
% Loss Variant B ($L_{b}$): Cosent loss is used specifically for the STS and Pair Classification tasks, while InfoNCE is used for the rest Retrieval tasks.
% Loss Variant C ($L_{c}$): Classification and Clustering datasets are incorporated, and we use the loss function described in Section \ref{losscls} for its fine-tuning.

As presented in Table \ref{tab:ablation}, $L_{b}$ outperforms $L_{a}$ notably in both pair classification and STS tasks. This finding is consistent with previous work \cite{cosent}, which claims that rank loss is often a superior choice for pairs with fine-grained labels. Finally, after reformatting clustering and classification datasets as contrastive triplets as we discussed in \ref{losscls}, $L_{c}$ achieves substantial improvements in both classification and clustering tasks.
\subsubsection{Large dimension} \label{sec:large_dim}
We also compare the impact of training embedding dimension on the final performance. The experimental results are presented in Table \ref{tab:dim_ablation}. Contrary to our expectations, expanding the dimension does not yield additional benefits to the embedding model.

\subsubsection{MRL training} \label{sec:mrl_abl}
We also examine the impact of employing MRL on the model's performance. The results presented in Table \ref{tab:mrl_ablation} indicate that whether we utilize MRL or not, there is minimal alteration in the model's performance. Yet it does demonstrate the effectiveness of MRL as it enables the support of flexible dimension length without sacrificing performance compared to single-dimensional training.

% Compared with directly constructing infoNCE for finetuning, here we do a simple ablation experiment to claim the effectiveness of hybrid loss tuning. In table \ref{tab:ablation}, for the $La$, we only use InfoNCE to finetune the dataset of the 6 tasks, for $Lb$, we use cosent loss for sts specifically and others InfoNCE. for $Lc$, we additionally incoporate classification and clustering datasets and use the loss we mentioned in \ref{losscls} for finetuning. In addition, we also demonstrated the huge improvements brought by incorporating clustering and classification datasets.

\begin{table}[t]
    \small
    \centering
    \resizebox{1.0\linewidth}{!}{%
	\begin{tabular}{l|cccccc|c}
		\toprule
		Training Dim. & Class. & Cluster. & Pair. & Rerank. & Retrieval & STS & Avg. \\
		\midrule
		1,024 & 74.42 & 62.2 & 90.02 & 69.86 & 74.21 & 63.34 & 70.79  \\
		1,792 & 74.59 & 62.17 & 90.24 & 70 & 74.36 & 63.5 & 70.95  \\
		3,072 & 74.49 & 62.22 & 90.42 & 69.83 & 74.3 & 63.42 & 70.87 \\

     \bottomrule
	\end{tabular}
}
\caption{Model's performance under different training embedding dimension.} \label{tab:dim_ablation}
\end{table}

\begin{table}[t]
    \small
    \centering
    \resizebox{1.0\linewidth}{!}{%
	\begin{tabular}{lc|cccccc|c}
		\toprule
		& Eval Dim & Class. & Cluster. & Pair. & Rerank. & Retrieval & STS & Avg. \\
		\midrule
		w/ MRL & 1,792 & 74.59 & 62.17 & 90.24 & 70 & 74.36 & 63.5 & 70.95  \\
        w/o MRL & 1,792 & 74.5 & 62.33 & 90.13 & 69.75 & 74.39 & 63.38 & 70.84  \\
     \bottomrule
	\end{tabular}
}
\caption{The impact of using MRL training.}\label{tab:mrl_ablation}
\end{table}

% \begin{table}[t]
%     \small
%     \centering
%     \resizebox{1.0\linewidth}{!}{%
% 	\begin{tabular}{lc|cccccc|c}
% 		\toprule
% 		& Dim & Class. & Cluster. & Pair. & Rerank. & Retrieval & STS & Avg. \\
% 		\midrule
%                & 3072 & - & 62.17 & 90.24 & 70.00 & 74.36 & 63.5 & -  \\
% 		   & 1792 & 74.34 & 62.32 & 90.15 & 69.82 & 74.41 & - & -  \\
%              & 1024 & - & - & - &- & - & - & -  \\
%      \bottomrule
% 	\end{tabular}
% }
% \caption{}\label{tab:dim_ablation}
% \end{table}

\section{Conclusion}
In this report, we present Piccolo2, the new state-of-the-art chinese embedding model. Piccolo2 mainly focuses on general downstream training by leveraging the effectiveness of multi-task hybrid loss. In addition, Piccolo2 also supports the use of vectors with flexible dimension lengths through MRL training. The latest information about the Piccolo2 model will be synchronized on Hugging Face: \href{https://huggingface.co/sensenova}{https://huggingface.co/sensenova}

\bibliographystyle{plainurl}
\bibliography{ref}

\newpage
\section*{Appendix} \label{sec:appendix}
\subsection*{Prompt for the retrieval data} \label{data_generation}

Referring to method~\cite{wang2023improving}, We adopted a two-stage approach to data generation. In the first phase, we let GPT4 randomly generate a large number of topic phrases in various domains to ensure the diversity of the data. The prompt used is shown in the Table~\ref{tab4}. In the second phase, we generate text pairs containing query, positive text and negative text based on the large number of topic phrases obtained from the previous generation through a specially designed prompt. In order to enrich the dataset, we generate text pairs 10 times for each topic phrase, and for each text pair, we set the corresponding random parameter in the prompt to make the text pairs obtained each time completely different. For example, for query generation, we randomly set its word count, clarity, etc., and randomly set the length and comprehensibility of positive text and negative text. It should be mentioned that for the definition of negative text, we require that the result generated by GPT4 should contain some useful information in the query, but its usefulness and comprehensiveness should not exceed that of the positive text. The prompt used is shown in the Table~\ref{tab5}.

\begin{table}[t]
    \caption{Example of a Prompt to Generate a Topic Phrase in Phase 1, 
    \{Retrieval\_tasks[x1]\} and \{Retrieval\_tasks[x2]\} are examples of randomized references, \{NUM\} indicates the number of generated topic phrases.} 
    \begin{tabularx}{\textwidth}{l}
        \hline
        "Brainstorm a list of potentially useful text retrieval tasks. Here are a few \\ examples for your reference:  \\
    - \{Retrieval\_tasks[x1]\} \\
    - \{Retrieval\_tasks[x2]\} \\
    Please adhere to the following guidelines: \\
    - Specify what the text is, and what the desired documents are. \\
    - Each retrieval task should cover a wide range of queries, and should not be \\ too specific. \\
    Your output must always be string, the string is a json dict start with \{ and \\ ends with \}, the key is `tasks', and the value is a list of strings only, with \\ about \{NUM\} elements, and each element corresponds to a distinct retrieval \\ task in one sentence. Do not explain yourself or output anything else. \\ Be creative!" \\  \hline
    \end{tabularx}
    \label{tab4}
\end{table}

\begin{table}[t]
    \caption{Example of a prompt to generate text pairs containing query, positive text and negative text in Phase 2, ``\{task\}" is the topic phrase generated in the first phase, $``\{query\_type\}" \in $ \{extremely long-tail, long-tail, common\}, $``\{query\_length\}"  \in $ \{less than 5 words, 5 to 15 words, at least 10 words\}, $``\{clarity\}"  \in $ \{clear, understandable with some effort, ambiguous\}, $``\{num\_words\}"  \in $ \{50, 100, 200, 300, 400, 500\}, $``\{difficulty\}"  \in $ \{high school, college, PhD\}.}  
    \begin{tabularx}{\textwidth}{l}
        \hline
        ``You have been assigned a retrieval task: \{task\} \\
    Your mission is to write one text retrieval example for this task in JSON \\ format. The JSON object must contain the following keys: \\
    - `user\_query': a string, a random user search query specified by the retrieval \\ task. \\
    - `positive\_document': a string, a relevant document for the user query. \\
    - `hard\_negative\_document': a string, a hard negative document that only \\ appears relevant to the query. \\
    Please adhere to the following guidelines: \\
    - The `user\_query' should be \{query\_type\}, \{query\_length\}, \{clarity\}, and \\ diverse in topic. \\
    - All documents must be created independent of the query. Avoid copying \\ the query verbatim. It’s acceptable \\
    if some parts of the 'positive-document' are not topically related to the \\ query. \\
    - All documents should be at least \{num\_words\} words long. \\
    - The 'hard\_negative\_document' contains some useful information, but it \\ should be less useful or comprehensive compared to the 'positive\_document'. \\
    - Both the query and documents should be in {language}. \\
    - Do not provide any explanation in any document on why it is relevant or \\ not relevant to the query. \\
    - Both the query and documents require \{difficulty\} level education to \\ understand. \\
    Your output must always be a JSON object only, do not explain yourself \\ or output anything else. Be creative!"  \\
    \hline
    \end{tabularx}
    \label{tab5}
\end{table}

\end{document}